\documentclass[journal=jacsat,manuscript=article]{achemso}

\setkeys{acs}{keywords = true}
\usepackage{graphicx}
\usepackage{dcolumn}
\usepackage{bm}
\usepackage[symbol]{footmisc}
\usepackage{soul}
\usepackage{xcolor}
\definecolor{black}{rgb}{0.0,0.0,0.0}
\usepackage{changes}
\definecolor{black}{rgb}{0.0,0.0,0.0}
\usepackage{ulem}

\usepackage[version=3]{mhchem} 
\usepackage[colorlinks,linkcolor=blue]{hyperref} 
\usepackage{url}

\author{Yurui Qu}
\email{yuruiqu@zju.edu.cn} 
\affiliation{Department of Physics, Massachusetts Institute of Technology, Cambridge, MA 02139, USA}
\alsoaffiliation{State Key Laboratory of Modern Optical Instrumentation, College of Optical Science and Engineering, Zhejiang University, Hangzhou 310027, China}
\author{Li Jing}
\email{ljing@mit.edu} 
\affiliation{Department of Physics, Massachusetts Institute of Technology, Cambridge, MA 02139, USA}
\author{Yichen Shen}
\affiliation{Department of Physics, Massachusetts Institute of Technology, Cambridge, MA 02139, USA}
\author{Min Qiu}
\affiliation{State Key Laboratory of Modern Optical Instrumentation, College of Optical Science and Engineering, Zhejiang University, Hangzhou 310027, China}
\alsoaffiliation{School of Engineering, Westlake University, 18 Shilongshan Road, Hangzhou 310024, China}
\alsoaffiliation{Institute of Advanced Technology, Westlake Institute for Advanced Study, Westlake University, 18 Shilongshan Road, Hangzhou 310024, China}
\author{Marin Solja\v{c}i\' c}
\affiliation{Department of Physics, Massachusetts Institute of Technology, Cambridge, MA 02139, USA}

\title[An \textsf{achemso} demo]
  {Migrating Knowledge between Physical Scenarios based on Artificial Neural Networks}

\abbreviations{IR,NMR,UV}

\setkeys{acs}{etalmode=truncate,articletitle = true, maxauthors=0}

\begin{document}



\begin{abstract}
Deep learning is known to be data-hungry, which hinders its application in many areas of science when datasets are small. Here, we propose to use transfer learning methods to migrate knowledge between different physical scenarios and significantly improve the prediction accuracy of artificial neural networks trained on a small dataset. This method can help reduce the demand for expensive data by making use of additional inexpensive data. First, we demonstrate that in predicting the transmission from multilayer photonic film, the relative error rate is reduced by 50.5\% (23.7\%) when the source data comes from 10-layer (8-layer) films and the target data comes from 8-layer (10-layer) films. Second, we show that the relative error rate is decreased by 19.7\% when knowledge is transferred between two very different physical scenarios: transmission from multilayer films and scattering from multilayer nanoparticles. Next, we propose a multi-task learning method to improve the performance of different physical scenarios simultaneously in which each task only has a small dataset. Finally, we demonstrate that the transfer learning framework truly discovers the common underlying physical rules instead of just performing a certain way of regularization.
\end{abstract}


Deep learning is a powerful machine learning algorithm that discovers representations of data with multiple levels of abstraction based on multiple processing layers \cite{lecun2015deep}. Recently, deep learning has received an explosion of interest because it continuously pushes the limit of traditional image recognition, machine translation, decision-making as well as many other applications \cite{hinton2012deep,krizhevsky2012imagenet,mnih2015human,silver2016mastering}. Meanwhile, deep learning is also penetrating into other disciplines such as drug design \cite{ma2015deep,gawehn2016deep}, genetics \cite{leung2014deep,xiong2015human}, material science \cite{ramprasad2017machine} and physics, including classification of complex phases of matter \cite{schoenholz2016structural,wang2016discovering}, electromagnetic inverse problem \cite{peurifoy2018nanophotonic,liu2018training}, nanostructure design \cite{malkiel2017deep,ma2018deep}, high-energy physics \cite{baldi2014searching} and quantum physics \cite{biamonte2017quantum,deng2017quantum,carleo2017solving}. One drawback is that deep learning is a data-hungry method and can only work well if fed with massive data. However, collecting a large amount of data is slow and expensive for many numerical simulations, such as bands of three-dimensional (3D) photonic crystals \cite{joannopoulos2011photonic}, and even much more difficult for experiments, since it might require fabricating tens of thousands of samples \cite{noda2000full} or doing tens of thousands of measurements \cite{zahavy2018deep}. Similar situations are also common in other scientific areas in which collecting a large amount of simulated or experimental data is difficult. Direct learning from a small dataset results in under-represented features, which leads to poor performance. There has yet to emerge a solution to improve deep learning performance for scientific problems with a small dataset.

Transfer learning has attracted growing interest in recent years because it can significantly improve the performance in the target task through the transfer of knowledge from the source task that has already been learned \cite{pan2010survey,bengio2012deep,thrun2012learning,torrey2010transfer}. Transfer learning is a useful method when the source dataset is large and inexpensive and target dataset is small and expensive. Jason Yosinski et al. demonstrated on ImageNet that transferred layers can improve classification accuracy by 2\% on a new task after substantial fine-tuning \cite{yosinski2014transferable}. Andrei A. Rusu et al. showed that learning from a simulation and transferring the knowledge to a real-world robot can solve the problem that training models on a real robot is too slow and expensive \cite{rusu2016sim}. However, unlike classic transfer learning that only cares about doing well on one particular target task or domain, another method called multi-task learning can do well on all related tasks which are trained simultaneously and benefit from each other. Multi-task learning has been used successfully across many applications such as natural language processing \cite{collobert2008unified}, speech recognition \cite{huang2013cross,deng2013new} and computer vision \cite{girshick2015fast}.

In this Letter, we propose a deep neural network architecture with transfer learning ability that can significantly improve the performance of physical problems even if their datasets are small.
There are two types of physical problems: one is the forward prediction problem shown in this work (the goal is to predict some physical properties given a specific physical system), and the other is the inverse design problem (the goal is to design a specific physical system based on some desired physical properties, such as the inverse design of a thin film device with specific reflection spectrum proposed by Dianjing et. al\cite{liu2018training}). In this paper, we focus on the forward prediction problems like transmission from multilayer films and scattering from nanoparticles. Although we focus here on certain particular photonic problems, the approach proposed can easily be generalized to many other scientific problems. The deep neural network with transfer learning is investigated in several cases: (1) the source and target data come from similar physical problems, for example, transmission from multilayer films with different number of geometric layers. The spectrum error continuously decreases as more layers of the neural network are transferred. Note that the former "layers" is used for photonic structures and the latter "layers" is used for weights and biases in neural networks. The relative error reduction is 50.5\% (23.7\%) when the source data comes from 10-layer (8-layer) film and the target data comes from 8-layer (10-layer) film; (2) The source and target data come from very different physical problems. The source data are scattering cross-sections from 8-layer core-shell nanoparticles and the target data are transmissivities from 8-layer films. To our surprise, the relative error rate still deceases by 19.7\% after transferring knowledge from the nanoparticle scattering problem to the multilayer film transmission problem; (3) Multiple tasks are 8-layer, 10-layer, 12-layer, 14-layer films. The performance of multi-task learning outperforms the direct learning trained only on a specific group of data. The neural network with transfer learning can significantly improve the performance of neural networks with only a small dataset, which could benefit the applications of deep learning in many physical problems and other fields.

\begin{figure}
\includegraphics[width=100mm]{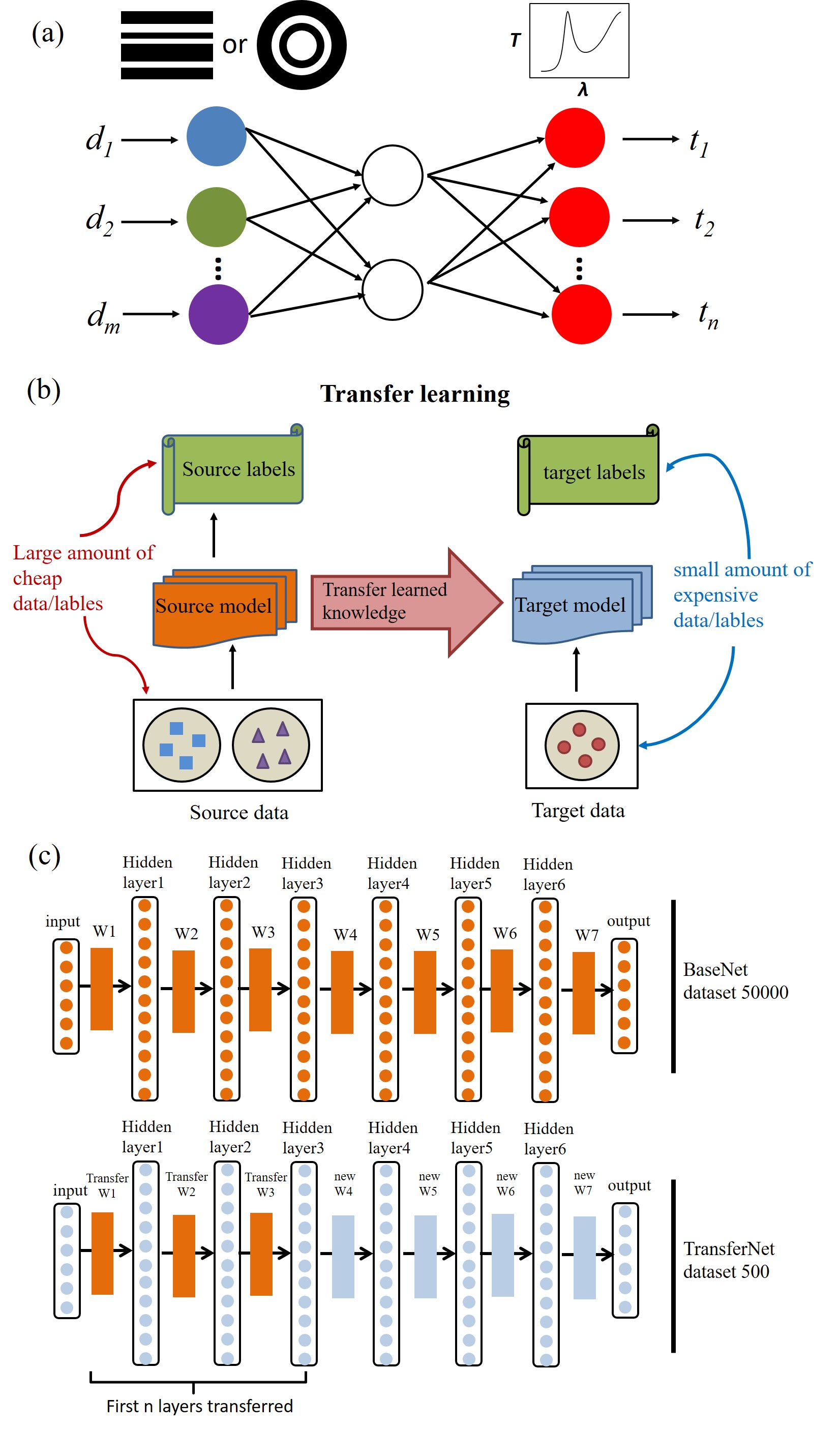}
\caption{(a) Illustration of the neural network architecture. The input is the thickness of each film or each shell of the nanoparticle, and the output is the transmission or scattering spectrum. Only one hidden layer is drawn here for convenience. Our actual neural network has six hidden layers. (b) Learning process of transfer learning. Transfer learning techniques can transfer the learned knowledge from the source model to the target model to improve the performance of the target task. Generally, the source domain has a large amount of inexpensive data while the target domain only has a small amount of expensive data. (c) Neural network structure of transfer learning. Top row: The base network (BaseNet) learns from scratch on a large source dataset of 50000 examples. Bottom row: The transfer network copies the first n layers from the BaseNet as the initialization weights and biases, and then the entire network is trained (fine-tuned) on the small target dataset of 500 examples.}
\label{fig1}
\end{figure} 

In classic deep learning, a model can be well trained for some task and domain if sufficient labeled data are provided. Let us assume that a task is the objective that our model aims to perform, e.g. predict the transmission spectra of 10-layer films, as shown in Fig. 1(a). We can now train a model on such a dataset and expect it to perform well on unseen data of a 10-layer film. However, this classic deep learning breaks down when we do not have sufficient labeled data for this task or domain. Training on a small dataset will cause collapse in performance because of severe overfitting problem. Transfer learning allows us to deal with this problem by leveraging the existing labeled data from some related task or domain, e.g. transmissivities from 8-layer films, or even a very different task like scattering cross-sections from core-shell nanoparticles. We try to store the knowledge gained in solving the source task in the source model and apply it to the target model to help the target task (see Fig. 1(b)).

Artificial neural network structure of our transfer learning method is shown in Fig. 1(c). The input data are the thicknesses of each film or shell (the materials were fixed), and the output data are the transmissivities sampled at points between 400 nm to 800 nm.  In order to further explore these and many other potential applications, we have publicly released all the data used in this paper \cite{Github}.
For different number of films as input, we use a one-dimensional mask (unnecessary position set as 0) to stretch the input into the same length. The thicknesses are between 30 nm to 70 nm, and materials are $SiO_2$ and $TiO_2$ for alternating layers of multilayer films and of core-shell nanoparticles.  We train a fully connected neural network called BaseNet on the source domain with a dataset of 50000 examples. 80\% of the data are used for training the network and the other 20\% are the validation data and the test data (10\% each), which are the same for the target task. More details about the neural network architecture have been included in supporting information.

TransferNet has the same network structure as BaseNet, while TransferNet copies the first n layers from the BaseNet as the initialization of weights and biases of the first n layers. The remaining higher layers of TransferNet are also initialized randomly with a normal distribution, and the entire TransferNet is fine-tuned simultaneously. TransferNet is trained on a target domain with a small dataset of 500 examples. The spectrum error decreases as the amount of training data increases, as shown in Fig. S1. There is a breaking point at around 1000 examples. When the amount of data is less than 1000, the error increases and the performance deteriorates sharply. We choose 500 examples in our training process to make sure it is truly a small amount of data. Next, we compare the transfer learning with the direct learning on the same dataset to demonstrate that transfer learning can truly improve the performance.

\begin{figure}
\includegraphics[width=100mm]{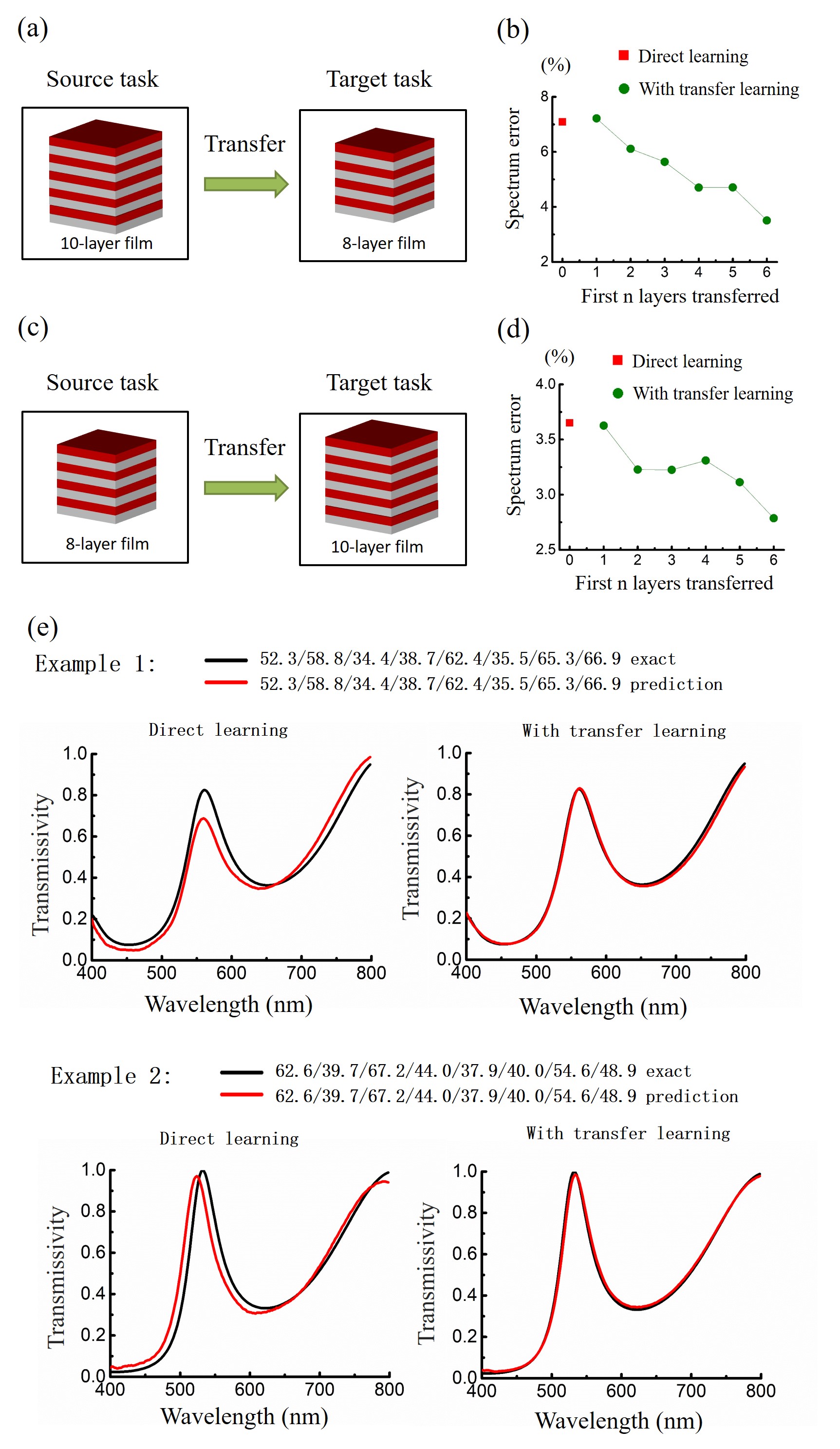}
\caption{(a) Illustration of transfer learning process and (b) test errors for the first n layers of BaseNet transferred when the source domain are 10-layer films and the target domain are 8-layer films. (c) and (d) are the case that the source domain are 8-layer films and the target domain are 10-layer films. (e) Two examples of transmission spectra for the case that the source domain are 8-layer films and the target domain are 10-layer films. Exact spectra are black lines and predicted spectra are red lines. Comparison of the direct learning to the transfer learning demonstrates that transfer learning can predict more accurate spectra than direct learning. 
}
\label{fig2}
\end{figure} 

The most general method of calculating the transmittance of a multilayer film is based on a matrix formula \cite{bass2010handbook} of the boundary conditions at the film surfaces derived from Maxwell’s equations (see Methods in Supporting Information). We use a neural network with direct learning and also with transfer learning, respectively, to approximate this transfer matrix formula, and we compare the performance in these two cases. We first explore the transfer learning between 8-layer films and 10-layer films, and try to transfer knowledge in both directions, as shown in Fig. 2(a) and (c). The spectrum error in this paper is defined as the average difference between the prediction and the exact result per spectrum point:
\begin{equation}
Error =\frac{1}{n}\sum_{i=1}^{n}\frac{\lvert{T_{prediction}(\lambda_i)-T_{exact}(\lambda_i)}\lvert}{T_{exact}(\lambda_i)}
\end{equation}
where n is the number of the spectrum points. In our case, we sampled between 400 nm to 800 nm with 2 nm step, so n=200 (not including 800 nm).

\begin{figure}
\includegraphics[width=150mm]{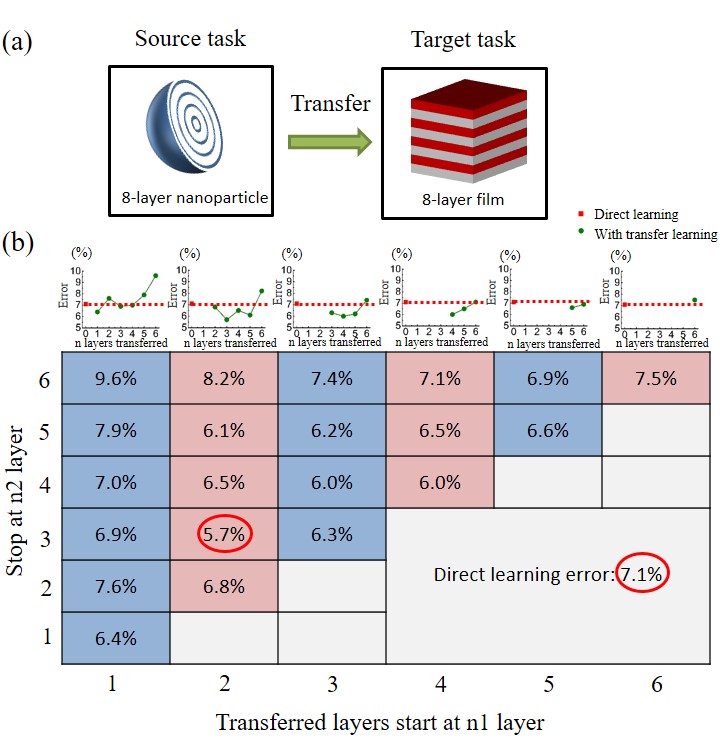}
\caption{(a) Illustration of the transfer learning process when the source domain are 8-layer nanoparticles and the target domain are 8-layer films. (b) Test errors when transferred layers begin at n1 layer and stop at n2 layer of trained BaseNet. The insets are spectra errors for each column.
}
\label{fig3}
\end{figure} 

 For target task of 8-layer film, direct learning has test error around 7.1\% (Fig. 2(b)). After transferring the first and second layers from trained BaseNet and retraining all the layers in TransferNet together, the spectrum error is reduced to 6.1\%. With more layers of the BaseNet transferred, the spectra errors continuously decrease. When 6 layers of BaseNet are transferred, the spectrum error decreases to about 3.5\%, which is 50.5\% relative reduction compared to direct learning. Transfer learning also works well when knowledge is transferred from 8-layer film to 10-layer film (Fig. 2(c)). Direct learning for 10-layer film has the spectrum error about 3.7\%. With 6 layers of BaseNet transferred, the spectrum error is reduced to around 2.8\% , which is 23.7\% relative error reduction (see Fig. 2(d)).

Two examples are presented in Fig. 2(e) to demonstrate that transfer learning can give a better prediction of the transmission spectrum compared to direct learning. Two examples come from the case that the source domain are 10-layer films and the target domain are 8-layer films. Black lines and red lines are theoretical and predicted transmission spectra, respectively. For the first example, the predicted spectrum using direct learning has lower peak transmissivity than the exact spectrum, and differences at other wavelengths are also obvious. For the second example, the entire predicted transmission spectrum based on direct learning shifts to shorter wavelength compared to the exact spectrum. However, the predicted spectra using transfer learning for both cases are much more accurate. This result is surprising because the spectra are predicted by the neural network which has only seen 400 training examples.

Next, we try to transfer knowledge between two very different tasks, the scattering from multilayer nanoparticles and the transmission from multilayer film, as shown in Fig. 3(a). Scattering cross-section from multilayer nanoparticle can be calculated using transfer matrix method, but in the forms of Bessel functions \cite{qiu2012optimization} (also see Methods in Supporting Information). More similar structures with different materials have been studied.\cite{abrashuly2019limits,sheverdin2019core} In Fig. 3(b), the first column of the table represents the spectrum error with the first n layers of the BaseNet transferred. The error of transfer learning is 0.7\% lower than direct learning (red dashed line) when only the first layer of the BaseNet is transferred before training the TransferNet. After transferring the first and second layers of the BaseNet, the error of transfer learning increases instead and surpasses the red dashed line, which is negative transfer that is harmful for learning the target task. Transferring the first 3 or 4 layers of the BaseNet can help to reduce the spectrum error a little lower than the direct learning. However, if the first 5 or 6 layers of the BaseNet are transferred together, the final performance will deteriorate sharply. From the results we can tell that some layers transferred from the BaseNet are specific to the nanoparticle scattering problem. These layers of the BaseNet will not help the target task learning at all, and can even be harmful for the final performance. Other layers of the BaseNet which are apparently general to both problems can be transferred and improve the target performance.

To find the layers of the BaseNet that contain the transferable physical knowledge, we utilize grid search to study the performance of transferring layers starting from $n1$ ending at $n2$, as shown in Fig. 3(b). The test error decreases to 5.7\% when $2^{nd}$ and $3^{rd}$ layers of the BaseNet are transferred, with around 19.7\% relative error reduction compared to direct learning. The results are  enlightening. Even for the case when there is not enough available data from a similar task, we can still utilize data from a different task, which largely expands the areas where this transfer learning method can be applied. Through this process, we are able to isolate the shared physical knowledge learned by the neural networks to some extent while keeping out the scenario specific knowledge. 

\begin{figure}
\includegraphics[width=100mm]{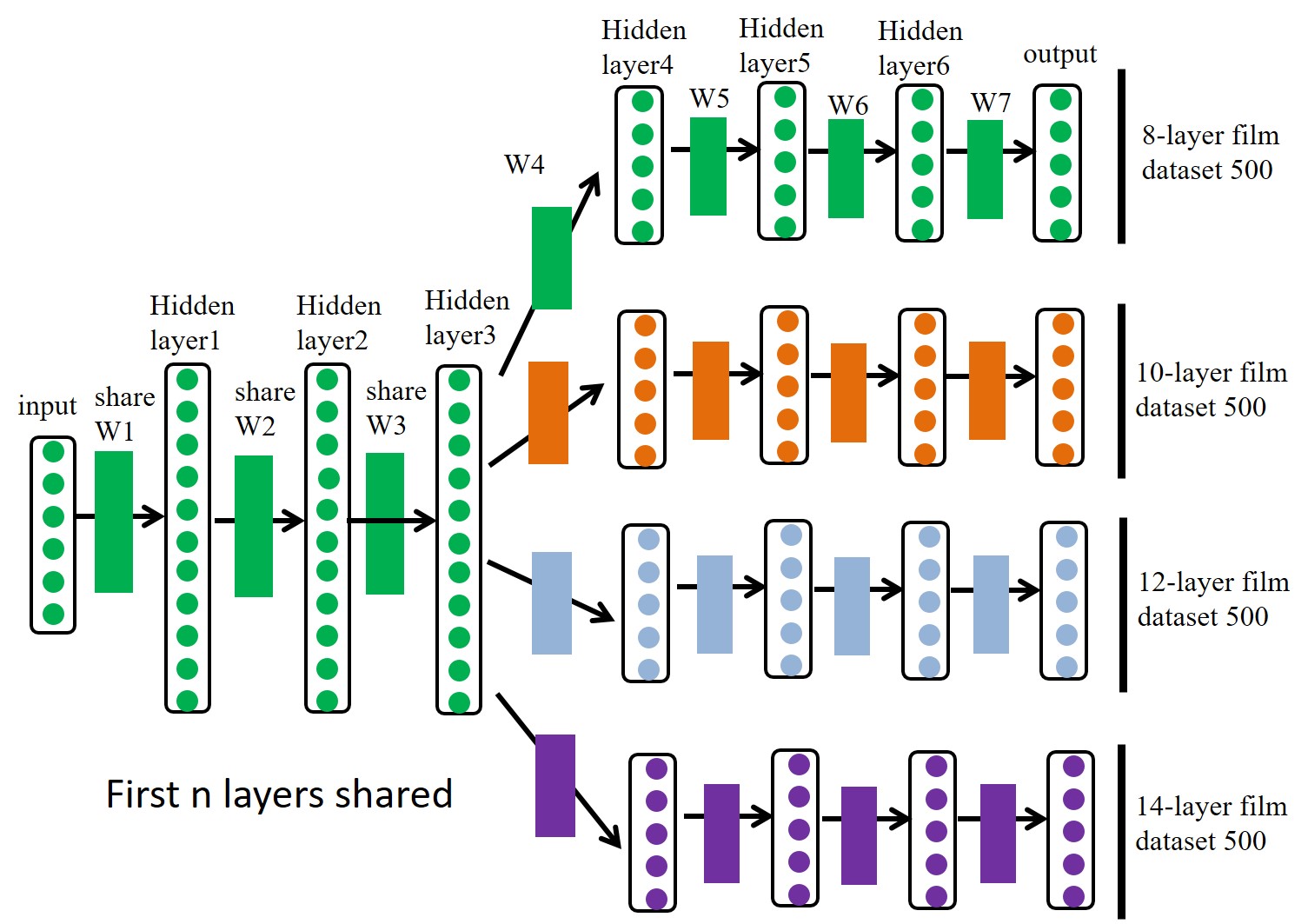}
\caption{ Neural network structure of multitask learning. The network shares the first n hidden layers and splits for the rest. Four target domains are 8-layer, 10-layer, 12-layer, 14-layer films, and each has a small dataset of 500 examples.
}
\label{fig4}
\end{figure} 

The transfer learning method described above requires a large amount of inexpensive data from related tasks. However, in some cases, there are several related tasks, but each of them has only a small amount of data. Classic transfer leaning method cannot work well in this case. Here, we introduce another knowledge transferring method called multi-task learning. As shown in Fig. 4 and Fig. S2, multi-task learning shares the first n hidden layers among all physical scenarios, while keeping several task-specific output layers. Here, four target data are from 8-layer, 10-layer, 12-layer, 14-layer films, and each has a small dataset of 500 examples. Each task can benefit from the knowledge learned in the other tasks, and the performance of each task can be improved compared to individual direct learning. This makes sense intuitively: the more tasks we are learning simultaneously, the more our model has to find a representation that captures all of the tasks and the less is our chance of overfitting the our original task. The key to this successful learning is to train the model for all the tasks simultaneously. All data are fed in each update of the model. The training algorithm needs to be adjusted slightly from the conventional backpropagation algorithm because of the split task-specific layers. When a training example is from 8-layer film, only the shared layers and the specific layers belonging to the 8-layer films task are updated. Other task-specific layers are kept intact. The same operation is done to train all four tasks.

We compare the spectrum errors of direct learning to that of multi-task learning in Table \ref{tab:multi-task}. Even if we only use four tasks learned together, each with a small dataset of 500 examples, multi-task learning has lower spectrum error than direct learning in each of four tasks. The relative reduction of the spectrum error is 9.9\%, 16.2\%, 1.7\% and 27.1\% for 8-layer, 10-layer, 12-layer, 14-layer films, respectively. We expect that the performance can be better with more target tasks. The best neural network structure is different for each target task, as shown in Fig. S3. For 8-layer and 10-layer films, the best performances are achieved when the first 2 hidden layers are shared. For 12-layer and 14-layer films, however, the best performances are achieved when the first 3 hidden layers are shared. We can also see that the performance deteriorates sharply if too many layers are shared. The reason is that the last several layers are specific for each task and cannot transfer knowledge among different tasks.

\begin{table}
\centering
\caption{Comparison of direct learning error to multi-task learning error}
\label{tab:multi-task}
\begin{tabular}{lcccc}
\hline\hline
Training dataset & 8 layers & 10 layers & 12 layers & 14 layers \\
\hline
Direct learning error & 7.1\% & 3.7\% & 6.0\% & 12.9\% \\
\hline
Multi-task learning error & 6.4\% & 3.1\% & 5.9\% & 9.4\% \\
\hline
Relative error reduction & 9.9\% & 16.2\% & 1.7\% & 27.1\% \\
\hline\hline
\end{tabular}
\end{table}

To demonstrate that the improved performance comes from transfer learning, not just from a certain way of regularization, we add L2 regularization, L1 regularization and dropout, several most widely used regularization methods, to the neural network of the direct learning and the transfer learning. The L2/L1 regularization can help the direct learning to improve a little, from 7.1\% error rate (without L2/L1 regularization) to 6.6\% (with L2/L1 regularization). However, the transfer learning with L2/L1 regularization always performs better than the direct learning with L2/L1 regularization, as shown in Fig. 5 (a) and (b). The lowest spectrum error of the direct learning with L2/ L1 regularization is 6.6\%; however, that of the transfer learning is 3.3\% with L2 regularization and 3.4\% with L1 regularization. This demonstrates that transfer leaning cannot be simply replaced by L2/L1 regularization.

Next, we add dropout to each hidden layer of the neural network. We control different keep\_prob to study the effect of the dropout on the neural network. When the keep\_prob equals 1.0, the case is the same as no dropout. As shown in Fig. 5 (c), dropout does not help to improve the performance of both direct learning and transfer learning, but makes it worse. These results demonstrate that the transfer learning framework truly discovers the common underlying physical rules instead of just performing a certain way of regularization.

\begin{figure}
\includegraphics[width=150mm]{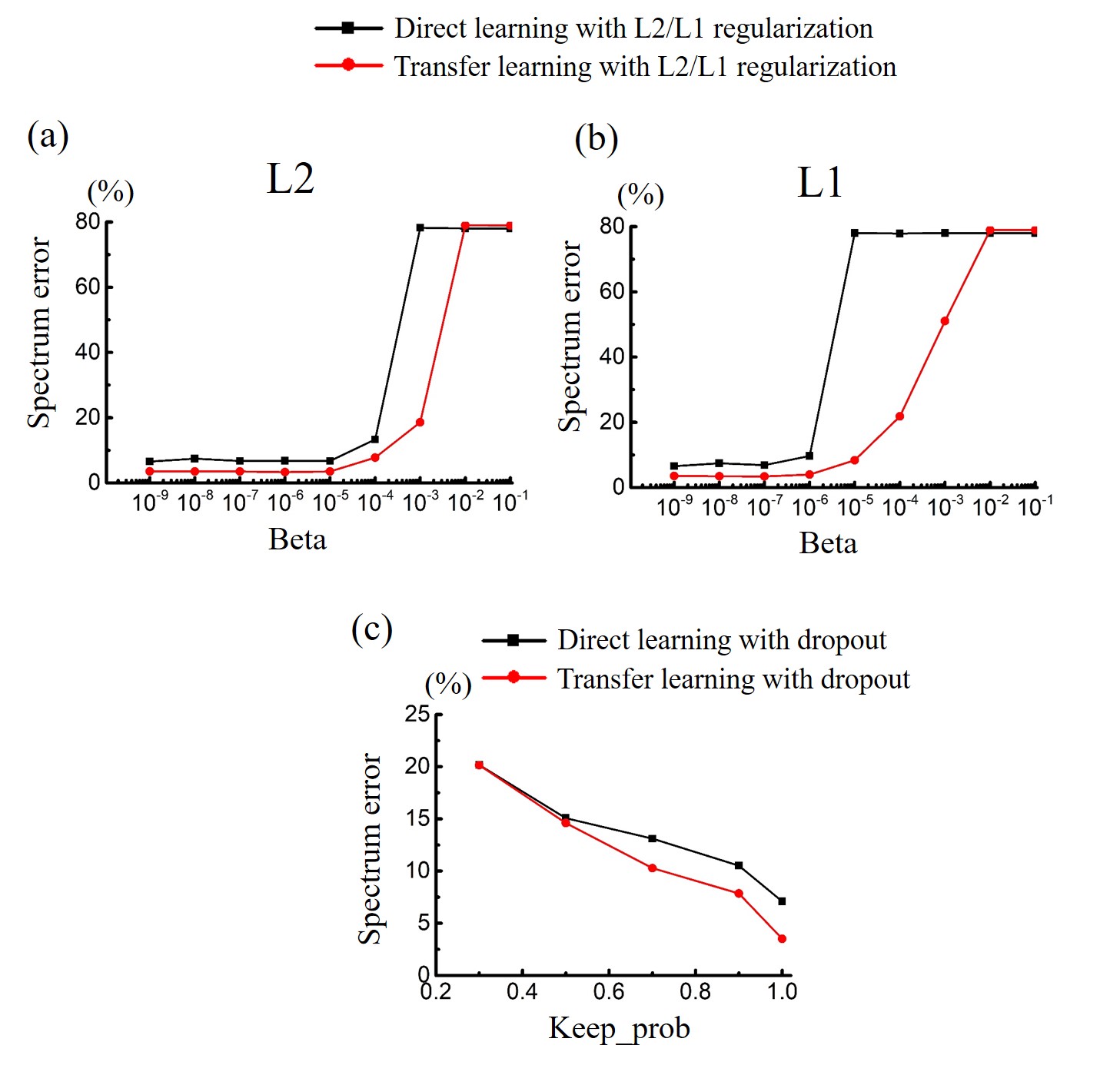}
\caption{ Spectrum errors for direct learning (black line) and transfer learning (red line) after adding (a) L2 regularization, (b) L1 regularization and (c) dropout to the neural network. The source domain are 10-layer films and the target domain are 8-layer films.
}
\label{fig5}
\end{figure} 

In conclusion, we present two transfer learning methods to help with the fact that deep learning methods cannot work well with small datasets in physical scenarios. We demonstrate that the neural network with transfer learning can give more accurate prediction compared to direct learning when trained on the same dataset. The knowledge in the neural network can be transferred not only between similar physical scenarios, such as transmission from multilayer films with different number of flims, but also between very different physical scenarios like scattering from core-shell nanoparticles and transmission from multilayer films. Multi-task learning, on the other hand, can improve the performance of several related tasks simultaneously even if each task only has a small dataset of 500 examples. The challenge of this transfer learning method is how to avoid negative transfer between two different tasks. Here we systematically select the general layers and specific layers in the neural network using grid search method, and it would be important to investigate if this process can be done automatically in the future. Looking forward, neural networks with transfer learning could not only benefit the development of deep learning in many physical problems of which datasets are expensive and small, but also in other areas of science such as biology, chemistry and material science.

\begin{acknowledgement}

We acknowledge discussions with John Peurifoy. Research was sponsored in part by the Army Research Office and was accomplished under Cooperative Agreement Number W911NF-18-2-0048. This material is based upon work supported in part by the National Science Foundation under Grant No. CCF-1640012.
This material is based upon work supported in part by the Semiconductor Research Corporation under Grant No. 2016-EP-2693-B. This material is also based in part upon work supported by the Defense Advanced 
Research Projects Agency (DARPA) under Agreement No. HR00111890042, as
well as in part by the MIT-SenseTime Alliance on Artificial Intelligence. This project was also supported by the National Key Research and Development Program of China (grant nos. 2017YFA0205700) and the National Natural Science Foundation of China (grant nos. 61425023). Yurui Qu was supported by Chinese Scholarship Council (CSC No. 201706320254).

\end{acknowledgement}

\begin{suppinfo}

Methods of analytically solving transmission from a multilayer film and scattering from multilayer nanosphere via the transfer matrix method; Technical details about the neural network architecture; Spectrum errors vs. the amount of data; Learning process of multi-task learning; Spectrum errors for multi-task learning; The training and validation errors over epochs; Spectrum errors when target task is 10-, 16- and 24-layer films.

\end{suppinfo}

\bibliography{achemso-demo}

\end{document}